\documentclass{article}
\usepackage{ijcai25}
\usepackage{amsmath}
\usepackage{amssymb}
\usepackage{times}
\usepackage{soul}
\usepackage{url}
\usepackage[hidelinks]{hyperref}
\usepackage[utf8]{inputenc}
\usepackage[small]{caption}
\usepackage{graphicx}
\usepackage{amsmath}
\usepackage{amsthm}
\usepackage{booktabs}
\usepackage{algorithm}
\usepackage{algorithmic}
\usepackage[switch]{lineno}
\usepackage{multirow}

\title{A Hybrid Multi-Factor Network with Dynamic Sequence Modeling for Early Warning of Intraoperative Hypotension}

\author{
Mingyue Cheng\textsuperscript{\rm 1},
Jintao Zhang\textsuperscript{\rm 1},
Zhiding Liu\textsuperscript{\rm 1},
Chunli Liu\textsuperscript{\rm 2}\footnote{Corresponding author}
\affiliations
\textsuperscript{\rm 1}State Key Laboratory of Cognitive Intelligence, University of Science and Technology of China\\
\textsuperscript{\rm 2}Intelligent Interconnected Systems Laboratory of Anhui Province, Hefei University of Technology\\
\emails
mycheng@ustc.edu.cn, \{zjttt, zhiding\}@mail.ustc.edu.cn, 
liuchunli@hfut.edu.cn
}

\begin{document}

\maketitle

\begin{abstract}
Intraoperative hypotension (IOH) prediction using past physiological signals is crucial, as IOH may lead to inadequate organ perfusion and significantly elevate the risk of severe complications and mortality. However, current methods often rely on static modeling, overlooking the complex temporal dependencies and the inherently non-stationary nature of physiological signals. We propose a Hybrid Multi-Factor (HMF) network that formulates IOH prediction as a dynamic sequence forecasting task, explicitly capturing both temporal dependencies and physiological non-stationarity. We represent signal dynamics as multivariate time series and decompose them into trend and seasonal components, enabling separate modeling of long-term and periodic variations. Each component is encoded with a patch-based Transformer to balance computational efficiency and feature representation. To address distributional drift from evolving signals, we introduce a symmetric normalization mechanism. Experiments on both public and real-world clinical datasets show that HMF significantly outperforms competitive baselines. We hope HMF offers new insights into IOH prediction and ultimately promotes safer surgical care. Our code is available at \url{https://github.com/Mingyue-Cheng/HMF}.

\end{abstract}

\section{Introduction}

\begin{figure}[t]
	\centering
	\includegraphics[width=0.47\textwidth]{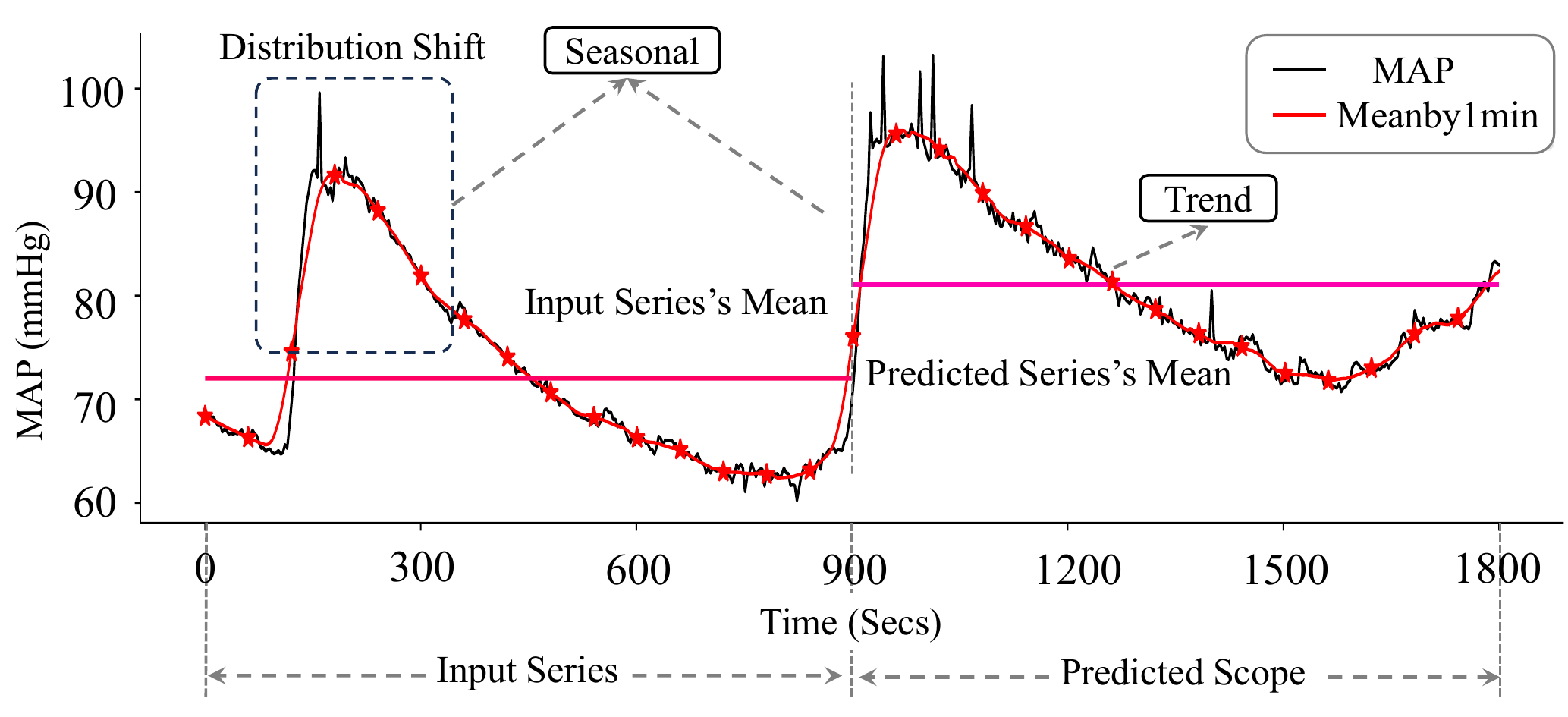} 
	\caption{Illustration of MAP sequence: temporal dynamics and distribution shifts in intraoperative hypotension events.}
	\label{fig2}
\end{figure}

Intraoperative mortality has decreased by a factor of $100$ over the past century, making deaths during surgery a rare occurrence~\cite{li2009epidemiology,xue2022perioperative}. However, mortality within the first month following surgery remains a significant concern, with approximately 2\% of patients undergoing inpatient noncardiac surgery dying within $30$ days postoperatively~\cite{spence2019association}—amounting to more than $4$ million deaths worldwide each year~\cite{saugel2021perioperative}. These postoperative deaths are most strongly linked to complications~\cite{naturedeath}, which are often triggered by intraoperative events. Among these, intraoperative hypotension (IOH) is a prevalent and clinically significant complication~\cite{kim2023intraoperative}, characterized by a sustained reduction in mean arterial pressure \footnote{MAP, derived from arterial blood pressure (ABP), is a key indicator of tissue perfusion. It is calculated as \( MAP = DBP + \frac{1}{3} (SBP - DBP) \)~\cite{MAPcaul}, where systolic blood pressure (SBP) and diastolic blood pressure (DBP) represent the maximum and minimum arterial pressures within a cardiac cycle, respectively.} (MAP).

Previous studies~\cite{hatib2018machine} have shown that IOH events can be predicted using machine learning applied to physiological signals. Clinical trials further indicate that early warnings and timely interventions help reduce hypotension severity and postoperative complications~\cite{hwang2023intraoperative,fernandes2021machine}. Recent advances primarily rely on handcrafted features combined with classifiers such as logistic regression and random forests~\cite{lee2021deep,davies2020ability}, but face two key limitations: (1) limited ability to capture temporal dependencies, and (2) lack of flexibility to handle complex real-world IOH scenarios. Additionally, fixed threshold-based IOH definitions may fail to reflect risk accurately in hypertensive or elderly patients, where more nuanced criteria are often needed.

To overcome these limitations, we reformulate IOH prediction as a multivariate time series forecasting problem. This formulation captures the temporal dynamics and evolving distributions inherent in physiological signals. Since hypotensive events are defined based on MAP series, accurate forecasting directly enhances risk identification. However, as illustrated in Figure~\ref{fig2}, several challenges arise. First, MAP waveforms comprise trends and periodic components, which may require separate modeling for improved accuracy. Second, redundancy in long MAP sequences increases computational cost and complicates representation learning. Third, sudden fluctuations alter statistical properties such as mean and standard deviation, making it difficult to maintain consistent performance. Addressing these challenges is essential for a robust and adaptive IOH prediction framework.

We propose the Hybrid Multi-Factor (HMF) framework, which explicitly models temporal structure and mitigates distribution shifts. HMF employs sequence decomposition to separate trend and seasonal components, yielding a more structured representation of historical dynamics. By incorporating multi-factor physiological signals such as MAP and SBP, HMF captures both short-term fluctuations and long-range dependencies. A patch-based Transformer encoder is adopted for efficient sequence representation, while a symmetric normalization mechanism compensates for evolving input distributions. Extensive experiments on public and private intraoperative datasets demonstrate that HMF consistently improves predictive performance and effectively models multi-factor temporal interactions. The key contributions are summarized as follows:
\begin{itemize}
	\item We reformulate IOH prediction as a multivariate time series forecasting problem and introduce HMF, which explicitly captures physiological signal dynamics.
	\item We highlight the challenges of IOH prediction based on historical physiological signal series, particularly the complexities in series structure and the non-stationary nature of physiological signals.
	\item We validate the effectiveness of the proposed approach through extensive experiments on two real-world datasets, demonstrating superior performance compared to existing methods.
\end{itemize}

\section{Related Work}
\subsection{Intraoperative Hypotension Forecasting}  
Existing methods for intraoperative hypotension (IOH) prediction predominantly adopt a static modeling approach, extracting handcrafted features from physiological signals rather than directly modeling their temporal dynamics. Early studies focused on high-fidelity arterial pressure waveforms, leading to the Hypotension Prediction Index (HPI)~\cite{R2}. Machine learning approaches, including ensemble methods~\cite{R5} and gradient boosting~\cite{R7}, integrated preoperative and intraoperative factors but largely treated data as independent samples, overlooking temporal dependencies. Deep learning models, such as RNN-based predictors~\cite{R8} and attention-based frameworks~\cite{R11}, improved performance by capturing sequential patterns but primarily relied on fixed-length feature representations, failing to fully model non-stationary hemodynamic changes. Recent interpretable models~\cite{R10} enhanced clinical applicability but remained feature-driven. Our work addresses these limitations by formulating IOH prediction as a multivariate time-series forecasting problem, explicitly capturing temporal dependencies and non-stationary dynamics in arterial blood pressure signals.

\subsection{Time Series and Sequence Forecasting}  
Time series forecasting is fundamental in domains such as finance, healthcare, and energy~\cite{cheng2025comprehensive,cheng2024convtimenet}. Traditional models like ARIMA~\cite{arima} and exponential smoothing struggle with high-dimensional physiological signals, while deep learning methods, including LSTMs~\cite{lstm} and GRUs~\cite{gru}, improve long-term dependency modeling.  Recent Transformer-based models enhance long-sequence forecasting effectiveness. Informer~\cite{informer} optimizes self-attention mechanisms through sparse attention and Autoformer~\cite{autoformer} leverages series decomposition to explicitly model trend and seasonal components, which enhances interpretability and improves forecasting stability over long horizons. Lightweight alternatives, such as DLinear~\cite{Dlinear}, demonstrate competitive performance by preserving temporal structures with simpler architectures. Due to the data sparsity issue~\cite{cheng2023timemae}, GPHT~\cite{GPHT} further introduces generative pretraining to improve forecasting transferability across datasets and horizons. While significant progress has been made in deterministic forecasting, probabilistic approaches such as diffusion-based models~\cite{csdi}~\cite{CDPM} are also gaining increasing attention for their ability to capture uncertainty. Complementarily, supervised architectures such as TimeDART~\cite{TimeDART} highlight the potential of discriminative learning in capturing temporal patterns effectively. On the other hand, the introduction of LLM-based models~\cite{TimeLLM}~\cite{GPTforTS} has enabled the integration of multimodal information. These advancements collectively provide a solid foundation for more accurate modeling of complex medical time series.

While existing models often struggle with real-time adaptability and the challenges posed by non-stationary physiological sequences in dynamic clinical settings, our work bridges this gap by combining efficient patch-based sequence encoding with structured modeling of physiological signals. This design enables rapid inference and robust adaptation to evolving patient states, thereby improving IOH prediction accuracy in dynamic clinical settings.

\begin{figure}[t]
	\centering
	\includegraphics[width=0.5\textwidth]{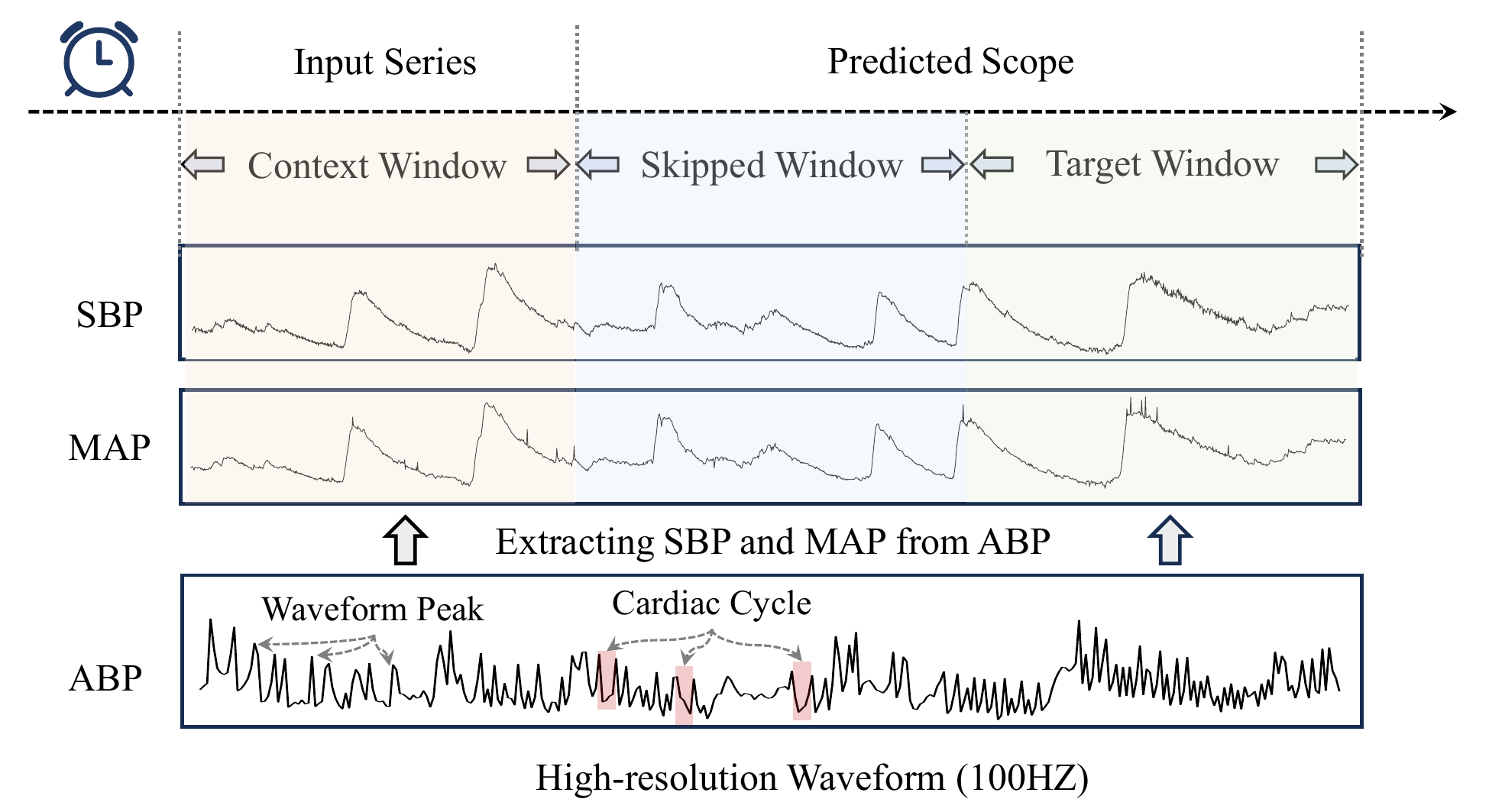} 
	\caption{Feature extraction and temporal window setup for blood pressure trend prediction: context, skipped, and target windows.}
	\label{fig:Problem}
\end{figure}

\begin{figure*}[h]
	\centering
	\includegraphics[width=0.8\textwidth]{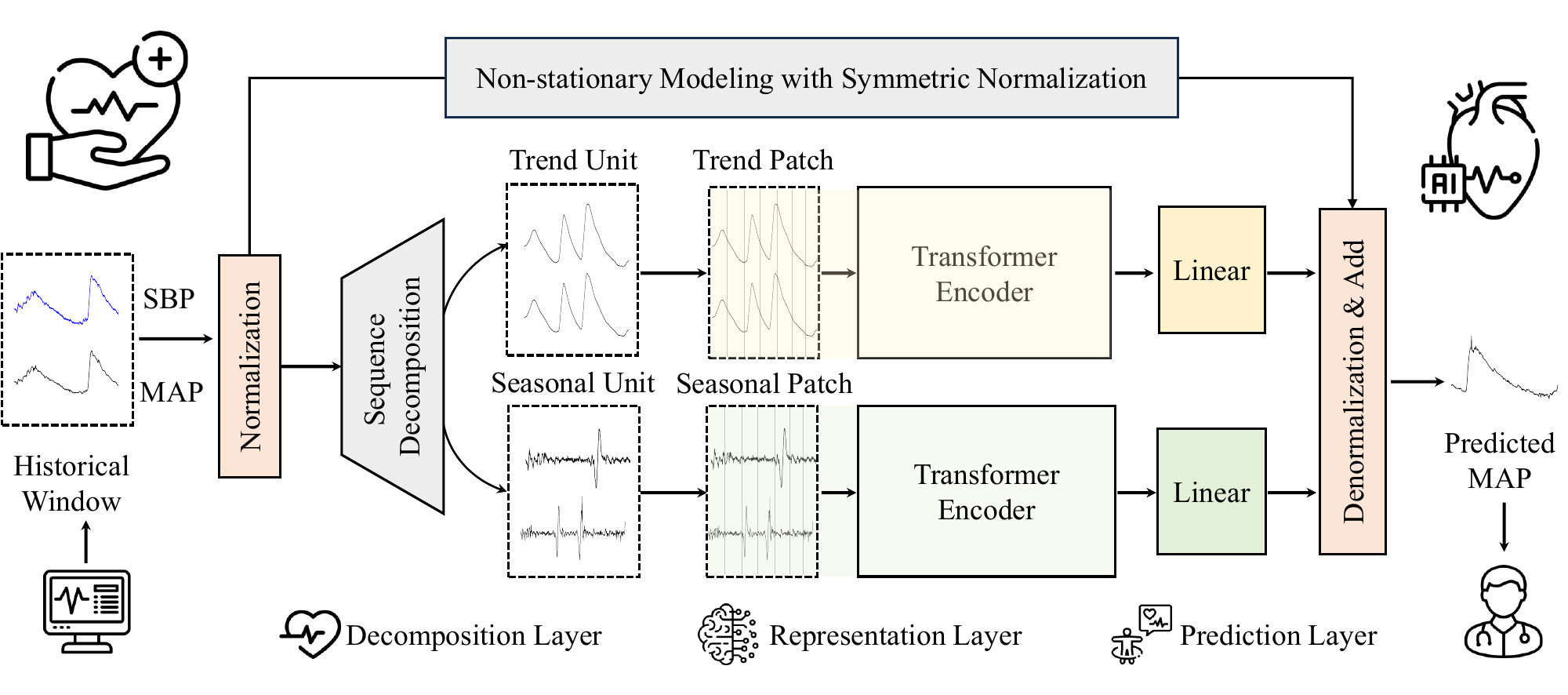} 
	\caption{Proposed hybrid multi-factor model for intraoperative hypotension prediction.}
	\label{fig:hmf}
\end{figure*}

\section{Preliminaries}

\subsection{Multi-factor Sequence Construction}  

To predict intraoperative hypotension (IOH), we analyze arterial blood pressure (ABP) dynamics to identify early indicators that precede hypotensive events, thereby enabling timely clinical intervention and risk mitigation. During surgery, ABP is continuously monitored at high frequency, capturing critical hemodynamic fluctuations. Prior studies have demonstrated a strong correlation between short-term variations in ABP and the onset of IOH.

Clinically, IOH is defined as a period when the mean arterial pressure (MAP) remains below 65 mmHg for at least one minute. Systolic blood pressure (SBP), the peak value in a cardiac cycle, is also considered due to its physiological relation to MAP. These two signals, MAP and SBP, serve as the primary predictive inputs for IOH forecasting. To construct reliable input sequences we extract clean ABP segments using a sliding window, remove artifacts and compute SBP and diastolic blood pressure (DBP), as shown in the Figure~\ref{fig:Problem}.

\subsection{Problem Definition}
Building on the multi-factor feature construction framework, we formulate IOH risk prediction as a dynamic sequence forecasting task. Given real-time ABP monitoring, the goal is to forecast future mean arterial pressure (MAP) trends to enable early warnings of hypotensive events. Let \( \mathbf{X}_{\text{MAP}} = \{x_{\text{MAP}}^1, \dots, x_{\text{MAP}}^T\} \) and \( \mathbf{X}_{\text{SBP}} = \{x_{\text{SBP}}^1, \dots, x_{\text{SBP}}^T\} \) represent historical MAP and systolic blood pressure (SBP) sequences over a context window of length \( T \). The objective is to predict MAP values \( \mathbf{\hat{Y}}_{\text{MAP}} = \{\hat{y}_{\text{MAP}}^{T+1}, \dots, \hat{y}_{\text{MAP}}^{T+\tau}\} \) over horizon \( \tau \). A hypotensive event is flagged if predicted MAP drops below a threshold \( \theta_{\text{MAP}} \). Further details on annotation and thresholding are provided in the supplementary material. By modeling multi-factor temporal dependencies, our approach improves IOH risk assessment and supports timely clinical decisions.

\section{The Proposed HMF}

\subsection{Framework Overview}  
The proposed Hybrid Multi-Factor (HMF) model predicts intraoperative hypotension (IOH) by leveraging MAP and SBP time series to capture both long-term trends and short-term fluctuations. As shown in Figure~\ref{fig:hmf}, the model consists of four key components: symmetric normalization, sequence decomposition, dynamic dependence modeling, and prediction.   First, symmetric normalization reduces inter-patient variability. Sequence decomposition separates trend and seasonal components to enhance feature learning. Transformer encoders process each component independently to capture temporal dependencies. Finally, the model fuses learned representations and reconstructs the MAP trajectory for forecasting.   By integrating multi-factor signals and sequence decomposition, the HMF model improves IOH prediction accuracy and adaptability to dynamic MAP variations.

\subsection{Symmetric Normalization}  

Non-stationarity in MAP and SBP series, driven by physiological variations during surgery, causes sudden shifts in statistical properties like mean and standard deviation~\cite{kim2021reversible}, challenging consistent IOH prediction. To mitigate these shifts and enhance reliability, we introduce a symmetric normalization module that ensures consistent feature scales across different surgical conditions, stabilizing model training and improving generalization in unseen cases.

Given a context window \( X \in \mathbb{R}^{L \times 2} \), instance normalization is applied as:  
\begin{equation}
	X' = \frac{X - \mu_X}{\sigma_X},
\end{equation}%
\noindent where \( \mu_X \) and \( \sigma_X \) are the mean and standard deviation of \( X \). This transformation allows the model to focus on relative changes rather than absolute values, reducing the impact of abrupt fluctuations. Additionally, by normalizing across instances, the method mitigates inter-patient variability, ensuring stable feature distributions.

\noindent After prediction, de-normalization restores the original scale:  
\begin{equation}
	\hat{Y} = \hat{Y}' \cdot \sigma_X + \mu_X,
\end{equation}%
\noindent where \( \hat{Y}' \) is the predicted sequence and \( \hat{Y} \) is the final output. This ensures stability while preserving raw characteristics, facilitating reliable and interpretable IOH forecasting.

\subsection{Decomposition Layer} 
The Arterial Blood Pressure(ABP) signal exhibits a complex waveform structure, consisting of multiple components like trend and periodic elements, which need to be individually modeled for accurate forecasting. After applying instance normalization, which results in the normalized sequence \( X' \), it becomes essential to decouple the physiological series into its trend and seasonal components. This decomposition allows for more precise modeling of each component, improving the overall forecasting accuracy. We follow the methodology of previous work\cite{NEURIPS2021_bcc0d400} to achieve this decoupling, enabling the model to better capture the inherent patterns in the ABP signal, is computed using:

\begin{align}
\text{Trend} = \text{AvgPool}(\text{Padding}(X')), 
\end{align}
\begin{align}
\text{Seasonal} = X' - \text{Trend},
\end{align}%
\noindent where \(\text{AvgPool}\) smooths the sequence by downsampling through averaging within a specified window, while \(\text{Padding}\) ensures full coverage across the entire sequence.

\subsection{Representation Layer}  
\paragraph{Patch Embedding.}  
To model the intricate dynamics of MAP and SBP series during surgery, we apply a patch embedding technique~\cite{patchtst} after the decomposition layer. This reduces sequence length, lowering computational complexity and improving efficiency. Patch-based modeling captures local patterns while mitigating noise and outliers, enhancing prediction robustness.  

The decomposed series \( W \in \mathbb{R}^{L \times 2} \) is transformed into a compact representation \( W_{\text{patch}} \in \mathbb{R}^{\frac{L}{S} \times d_{\text{model}}} \) using three 1D convolutional layers:  

\begin{equation}
	W_{\text{patch}} = \text{Conv1D}(W),
\end{equation}%
where \(\text{Conv1D}\) applies three consecutive 1D convolutions to \( W \). The embedding dimension \( d_{\text{model}} \) is chosen to extract key features, improving IOH prediction.  

To encode temporal structures, we introduce learnable positional encodings:  
\begin{equation}
	W_{\text{pos}} = W_{\text{patch}} + \text{PositionalEncoding}(W_{\text{patch}}).
\end{equation}%

These encodings help preserve temporal dependencies, crucial for MAP forecasting and IOH detection, where event timing directly impacts prediction accuracy.

\paragraph{Sequence Dependence Modeling.}  

Capturing cross channel dependencies is essential for accurate prediction of multivariate physiological time series~\cite{han2024capacity}. Effective IOH prediction requires capturing both short-term and long-term dependencies in MAP and SBP series. The Transformer encoder, with its self-attention mechanism, is well-suited for this task, dynamically weighting the importance of different time segments to model complex temporal patterns.  

After obtaining patch embeddings \( X_{\text{pos}} \) via patch embedding and positional encoding, the Transformer encoder processes them through self-attention and feedforward layers to learn sequence dependencies. Given positional embeddings \( X_{\text{pos}} \in \mathbb{R}^{\frac{L}{S} \times d_{\text{model}}} \), the dependencies are computed as:  

\begin{equation}
	Z = \text{TransformerEncoder}(W_{\text{pos}}),
\end{equation}%

\noindent where \( Z \in \mathbb{R}^{\frac{L}{S} \times d_{\text{model}}} \) encodes cross-segment relationships in MAP and SBP. This formulation ensures that the model captures subtle fluctuations and interactions essential for dynamic sequence modeling. By leveraging self-attention mechanism, the Transformer enhances predictive capability, aligning with the multi-factor feature learning framework to improve IOH forecasting.

\begin{table*}[t]
	\centering
        \small
	\label{table:dataset_details}
	\resizebox{0.85\linewidth}{!}{%
		\begin{tabular}{ccccccc}
			\hline
			\textbf{Dataset} & \textbf{Patient Number} & \textbf{Sampling Rate (s)} & \textbf{Predicted Scope} & \textbf{Training Set} & \textbf{Validation Set} & \textbf{Testing Set} \\
			\hline
			\multirow{6}{*}{CH-OPBP} & \multirow{6}{*}{1,083} & \multirow{3}{*}{1} & 300 & 442,858 & 54,406 & 58,019 \\
			& & & 600 & 489,273 & 58,810 & 66,071 \\
			& & & 900 & 519,339 & 61,776 & 71,130 \\
			\cline{3-7}
			& & \multirow{3}{*}{3} & 100 & 148,274 & 18,243 & 19,401 \\
			& & & 200 & 164,049 & 19,719 & 22,124 \\
			& & & 300 & 174,311 & 20,764 & 23,848 \\
			\hline
			\multirow{3}{*}{VitalDB} & \multirow{3}{*}{1,522} & \multirow{3}{*}{3} & 100 & 466,402 & 61,158 & 58,042 \\
			& & & 200 & 589,414 & 78,037 & 74,094 \\
			& & & 300 & 689,650 & 91,902 & 87,685 \\
			\hline
		\end{tabular}
	}
		\caption{Statistics of training, validation, and testing sets across different datasets and prediction scopes.}
        \label{table:dataset_details}
\end{table*}

\subsection{Prediction Layer}  
To ensure the temporal patterns captured by the Transformer encoder are effectively utilized in MAP predictions, we employ a linear layer to map the sequence representation \( Z \) to the predicted output. This transformation allows the model to preserve sequence dependencies while translating extracted features into precise forecasts. To maintain temporal coherence, we use a recurrent forecasting strategy where each predicted step conditions the next, ensuring consistency across the forecasted sequence. The mapping from \( Z \in \mathbb{R}^{\frac{L}{S} \times d_{\text{model}}} \) to \( Y \in \mathbb{R}^{\frac{L}{S} \times \frac{2TS}{L}} \) is formulated as:

\begin{equation}
	Y = Z \cdot W + b,
\end{equation}%
\noindent where \( W \) and \( b \) denote the transformation matrix and bias vector, respectively. Linear transformations have been widely adopted in sequence forecasting~\cite{Dlinear}, facilitating the direct conversion of encoded representations into accurate future MAP predictions.

\subsection{Optimization Strategies}  
To improve prediction accuracy, we integrate a patch-based autoregressive method into our model. This approach begins with the final patch of decomposed component \( W \), denoted as \( Y_{0} = W_{\frac{L}{S}} \), and sequentially generates subsequent patches using the autoregressive formula:

\begin{equation}
	P(Y'_{i+1}) = \prod_{j=1}^{i} P(Y'_{j+1} | Y'_{j}),
\end{equation}%
\noindent in which \( Y'_{i+1} \) represents the next patch. The predicted sequence can be represented as:

\begin{equation}
	\hat{Y}' = \{ Y'_{1}, Y'_{2}, \dots, Y'_{\frac{T}{S}} \}.
\end{equation}%
By leveraging prior patches, this method improves temporal dependency modeling for MAP forecasting.   The model is optimized by minimizing the mean squared error (MSE) between the predicted and actual sequences:

\begin{equation}
	\text{MSE} = \frac{1}{T} \sum_{t=1}^{T} \left( Y_t - \hat{Y}_t \right)^2.
\end{equation}%
This optimization refines forecasting accuracy, enhancing IOH prediction effectiveness.

\begin{figure}[h!]
	\centering    
	\includegraphics[width=\linewidth]{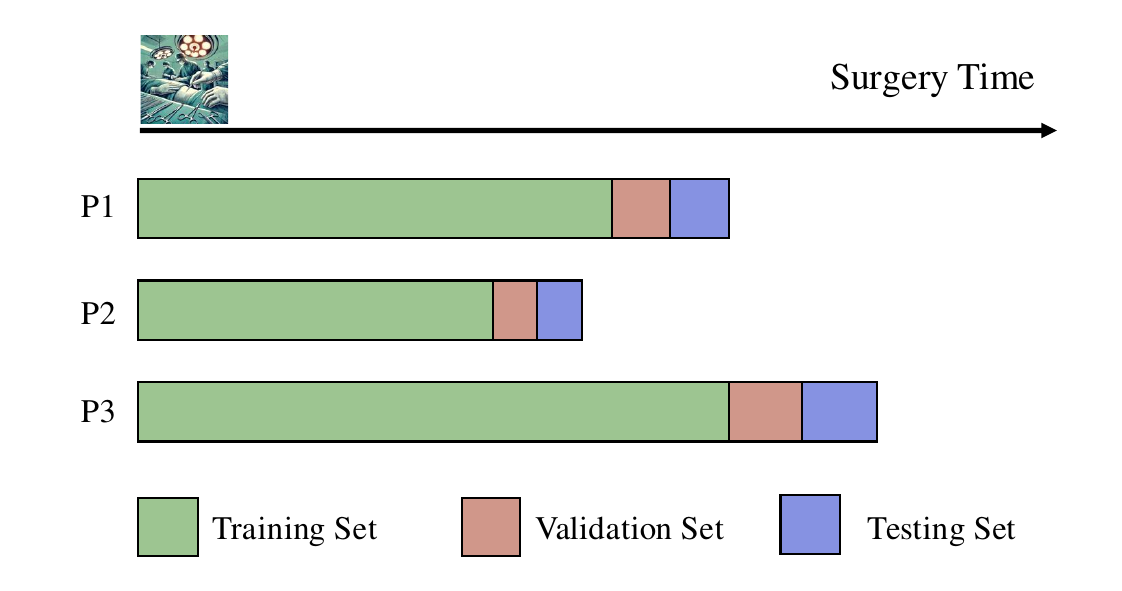}
	\caption{Visualization of data partitioning across different patients during surgery time.}
	\label{fig:patient_data_partitioning}
\end{figure}

\section{Experiments}

\subsection{Experimental Setup}

\paragraph{Dataset Description.}  

We conduct experiments on two real-world clinical datasets: \textbf{CH-OPBP} and \textbf{VitalDB}~\cite{lee2018vital}. The \textbf{CH-OPBP} dataset is collected from real-world intraoperative monitoring records of 3,422 patients, sampled continuously at 100Hz. The data acquisition process strictly adheres to ethical guidelines and protocols approved by the hospital ethics committee, ensuring full compliance with patient privacy regulations and relevant clinical research standards. After resampling to 1-second and 3-second intervals and filtering out records shorter than one hour, a total of 1,083 valid records remain. The data acquisition process also strictly follows patient de-identification protocols and has been formally approved by the Institutional Review Board of Anhui Provincial Hospital. Data collection spans the period from February 27, 2023, to August 4, 2023. The \textbf{VitalDB} dataset initially includes 6,388 intraoperative records, resampled uniformly at 3-second intervals. After excluding records with more than 20\% missing data, 1,522 records are retained for use. To ensure reliable model evaluation, both datasets adopt an 80\%-10\%-10\% split for training, validation, and testing, while preserving temporal consistency. The CH-OPBP dataset supports multiple prediction horizons to evaluate forecasting flexibility, whereas the VitalDB dataset focuses specifically on 3-second resolution data. Table~\ref{table:dataset_details} summarizes data partitioning across different sampling rates and prediction scopes. Figure~\ref{fig:patient_data_partitioning} visualizes the partitioning strategy, segmenting each patient's surgical timeline for structured model training and evaluation.

\begin{algorithm}
	\caption{IOH Events Detection}
	\label{alg3}
	\textbf{Input}: $O_{\text{pred}}$: predicted sequence of MAP values, $t$: minimum duration of IOH, $T_{\text{instance}}$: length of the target window, $L_{\text{actual}}$: actual labels sequence.\\
	\textbf{Output}: $J_{\text{actual}}$: boolean indicating the presence of an IOH event in the actual instance, $J_{\text{prediction}}$: boolean indicating the presence of an IOH event in the predicted instance.
	\begin{algorithmic}[1] 
		\STATE $J_{\text{actual}} = 0$
		\STATE $J_{\text{prediction}} = 0$
		\FOR{$i$ in range(0, $T_{\text{instance}}-t+1$)}
		\STATE $\text{sum}_{\text{actual}} = \text{sum}(L_{\text{actual}}[i : i + t])$
		\IF {$\text{sum}_{\text{actual}} > 0$}
		\STATE $J_{\text{actual}} = 1$
		\ENDIF
		\STATE $\text{sum}_{\text{prediction}} = \sum_{j=0}^{t-1} (O_{\text{pred}}[i+j] \leq 65)$
		\IF {$\text{sum}_{\text{prediction}} > 0.8*t$}
		\STATE $J_{\text{prediction}} = 1$
		\ENDIF
		\ENDFOR
		\STATE \textbf{return} $J_{\text{actual}}$, $J_{\text{prediction}}$
	\end{algorithmic}
\end{algorithm}

\begin{table*}[h!]
	\small
	\centering
	\captionsetup{justification=centering}
	\begin{tabular}{l c l c c c c c}
		\hline
		\textbf{Datasets} & \textbf{Sample (s)} & \textbf{Model} & \textbf{MSE} & \textbf{MAE} & \textbf{AUC} & \textbf{Accuracy (\%)} & \textbf{Recall (\%)} \\
		\hline
		\multirow{7}{*}{CH-OPBP} & \multirow{7}{*}{1} & Arima & 130.2702 & 8.8526 & 0.5963 & \textbf{77.32} & 24.00 \\
		& & LR & \textemdash & \textemdash & 0.5054 & 76.08 & 35.10 \\
		& & LSTM & 118.1246 & 9.0985 & 0.5295 & 76.78 & 6.58 \\
		& & Transformer & 126.7972 & 9.3809 & 0.5919 & 75.63 & 22.43 \\
		& & Informer & 103.7028 & 8.0757 & 0.6452 & 72.26 & 35.38 \\
		& & DLinear & 125.6786 & 9.3232 & 0.5331 & 71.60 & 7.54 \\
		& & HMF & \textbf{93.2677} & \textbf{7.5823} & \textbf{0.7352} & 75.29 & \textbf{67.98} \\
		\hline
		\multirow{7}{*}{CH-OPBP} & \multirow{7}{*}{3} & Arima & 112.9281 & 8.1606 & 0.5928 & 75.70 & 25.65 \\
		& & LR & \textemdash & \textemdash & 0.6774 & 75.71 & 54.49 \\
		& & LSTM & 124.4213 & 9.8814 & 0.5000 & 75.53 & 0.00 \\
		& & Transformer & 104.8545 & 8.3441 & 0.5970 & 74.44 & 23.12 \\
		& & Informer & 111.0393 & 8.3883 & 0.6278 & \textbf{79.81} & 30.37 \\
		& & DLinear & 123.8899 & 9.2951 & 0.5413 & 74.10 & 11.98 \\
		& & HMF & \textbf{86.4927} & \textbf{7.2828} & \textbf{0.7413} & 61.38 & \textbf{70.13} \\
		\hline
		\multirow{7}{*}{VitalDB} & \multirow{7}{*}{3} & Arima & 257.3701 & 13.1127 & 0.5250 & 59.31 & 8.53 \\
		& & LR & \textemdash & \textemdash & 0.5595 & 62.60 & 33.47 \\
		& & LSTM & 188.7123 & 11.8613 & 0.5000 & \textbf{75.62} & 0.00 \\
		& & Transformer & \textbf{158.7031} & 10.6901 & 0.5040 & 73.51 & 0.93 \\
		& & Informer & 158.7873 & 10.8987 & 0.5003 & 75.01 & 0.05 \\
		& & DLinear & 175.1144 & 11.4968 & 0.5074 & 65.09 & 1.86 \\
		& & HMF & 165.7575 & \textbf{9.3845} & \textbf{0.6468} & 69.27 & \textbf{45.87} \\
		\hline
	\end{tabular}
    \caption{%
		Performance comparison of IOH prediction between our HMF and baseline models on two datasets .}
	\label{table:comparison_result}
\end{table*}

\paragraph{Compared Baselines.}  
To evaluate HMF, we compare it with traditional and deep learning-based forecasting models. For traditional methods, we use ARIMA~\cite{arima} and Logistic Regression. ARIMA, a statistical time series model, is trained on 0.5\% of the test data for efficiency. Logistic Regression, a standard classifier, is enhanced with 1,566 features extracted via tsfresh\footnote{\url{https://github.com/blue-yonder/tsfresh}} and evaluated on 1\% of the dataset.  For deep learning baselines, we include LSTM~\cite{lstm}, Transformer~\cite{transformer}, Informer~\cite{informer}, and DLinear~\cite{Dlinear}. LSTM captures sequential dependencies via recurrence, while Transformer leverages self-attention for long-range interactions. Informer improves efficiency with sparse attention, making it suitable for long-sequence forecasting. DLinear, a lightweight alternative, models trend and seasonal components separately. Further implementation details and hyperparameter settings are in the supplementary material.

\paragraph{Implementation and Evaluation Details.}  
The HMF model predicts future MAP values based on physiological forecasts of both MAP and SBP, utilizing a 15-minute context window as input. Predictions extend over 5, 10, and 15 minutes to capture short- and long-term trends. Model training minimizes Mean Squared Error (MSE), with evaluation using both MSE and Mean Absolute Error (MAE), focusing on hypotensive segments (\( L_{\text{actual}} = 1 \)).   Although both MAP and SBP series are predicted, IOH assessment relies solely on the forecasted MAP trajectory. Following clinical guidelines~\cite{intraoperative}, a hypotensive event is identified when predicted MAP remains below \( \theta_{\text{MAP}} = 65 \) mmHg for at least 1 minute. To enable effective early warning and facilitate clinical decision-making, a 2-minute skip window~\cite{wijnberge} is applied. Performance is assessed using accuracy, recall, and Area Under the Curve (AUC), with AUC serving as the primary metric for evaluating the model’s ability to differentiate hypotensive trends. Algorithm~\ref{alg3} details the IOH event detection process by comparing predicted and actual MAP sequences in a time-series forecasting framework. Further methodological details, including ground truth labeling and evaluation protocols, are provided in supplement material part.

\begin{table*}[h!]
	\small
	\centering
	\begin{tabular}{clccccc}
		\hline
		Dataset & Model & MSE & MAE & AUC & Accuracy (\%) & Recall (\%) \\ \hline
		\multirow{3}{*}{CH-OPBP} 
		& HMF (Our model) & \textbf{86.4927} & \textbf{7.2828} & \textbf{0.7413} & 75.53 & \textbf{70.13} \\
		& w/o instance normalization  & 106.9671        & 8.5005          & 0.5891          & 77.72 & 20.73 \\
		& w/o sequence decomposition & 105.7231        & 8.4964   & 0.5750  & \textbf{78.10} & 17.50 \\ 
		\hline
	\end{tabular}
	\caption{Ablation study on instance normalization and sequence decomposition in HMF on the CH-OPBP dataset.}
    \label{table:performance_comparison}
\end{table*}

\subsection{Experimental Results}
\paragraph{Main Results Analysis.}  
Table~\ref{table:comparison_result} presents a comprehensive evaluation of predictive models for dynamic IOH prediction, highlighting the advantages of the HMF model.  The results indicate that HMF consistently outperforms baseline methods across datasets and sampling rates. Its ability to capture the intricate dynamics of blood pressure trends is reflected in improved predictive accuracy and classification performance. Notably, HMF demonstrates robustness in handling the non-stationary and complex nature of intraoperative blood pressure data, where traditional models often struggle.  In the IOH prediction task, HMF shows strong performance in detecting hypotensive events. Its improvements across different sampling rates suggest adaptability to varying temporal granularities, a crucial factor in clinical applications. Compared to LSTM and Transformer, which perform well in some scenarios but face challenges with long sequence dependencies, HMF effectively models temporal dependencies and coupling effects. Informer and DLinear, despite their advantages in general forecasting tasks, show limitations in handling intraoperative data. Overall, HMF maintains stable performance across diverse conditions, demonstrating its potential to enhance IOH prediction in complex clinical settings.

\paragraph{Ablation Study of Key Components.}  
To evaluate the impact of key components in the HMF framework, we conduct an ablation study on the CH-OPBP dataset. Table~\ref{table:performance_comparison} presents the results, highlighting the roles of instance normalization and sequence decomposition in MAP forecasting and IOH detection. The full HMF model, integrating both components, achieves the best overall performance, suggesting their complementary effects in enhancing predictive capability.   Removing instance normalization leads to a notable performance drop, indicating its importance in managing the non-stationary nature of intraoperative blood pressure data. Without normalization, the model struggles with variability, affecting its ability to detect IOH events consistently. Similarly, excluding sequence decomposition degrades performance, suggesting its role in capturing complex temporal dependencies. Without decomposition, the model may fail to identify underlying trends, limiting predictive accuracy.  These findings underscore the importance of both components in improving model accuracy. Integrating instance normalization and sequence decomposition enhances generalization, leading to more  effective IOH predictions.

\begin{table}[h!]
	\small
	\centering
	\begin{tabular}{lcccc}
		\hline
		Methods & Patient ID & MSE & MAE & AUC \\ \hline
		\multirow{2}{*}{Transfer} & 1 & 37.3299 & 5.6241 & 0.5375 \\
		& 2 & 39.9698 & 6.2213 & 0.6100 \\ \hline
		\multirow{2}{*}{Non-Transfer} & 1 & 18.6823 & 3.8187 & 0.2375 \\
		& 2 & 47.6620 & 6.8317 & 0.5600 \\ \hline
	\end{tabular}
	\caption{Comparison of transfer and non-transfer learning on individual patients.}
	\label{table:transfer_person}
\end{table}

\begin{table}[h!]
	\small
	\centering
	\begin{tabular}{lcccc}
		\hline
		Methods& Feature & MSE & MAE & AUC \\ \hline
		\multirow{2}{*}{Transfer} 
		& Elderly   & 118.9462 & 9.1424 & 0.6089 \\
		& Young & 38.0568  & 5.0047 & 0.8040 \\ \hline
		\multirow{2}{*}{Non-Transfer} 
		& Elderly   & 66.5058  & 6.5874 & 0.7850 \\
		& Young & 44.1721  & 5.2447 & 0.8540 \\ \hline
	\end{tabular}
	\caption{Comparison of transfer and non-transfer learning on elderly and young subgroups.}
	\label{table:transfer_feature}
\end{table}

\paragraph{Performance Analysis under Transfer Learning Settings.}  
Assessing model transferability across patients and demographic groups is essential for IOH prediction. Transfer learning enhances the robustness and generalizability of the HMF framework. To evaluate this, we apply the HMF model, trained on CH-OPBP, to new patient cohorts and age groups, comparing performance with and without transfer learning. Table~\ref{table:transfer_person} presents cross-patient transfer results, where a model trained on one patient is applied to another. AUC scores suggest that transfer learning preserves or enhances predictive performance, indicating the model's ability to capture shared physiological patterns. This reduces reliance on patient-specific data while maintaining accuracy, though individual training may still offer advantages in distinct cases. Table~\ref{table:transfer_feature} presents cross-group results between age cohorts. The non-transfer model performs better in elderly patients, suggesting that model better capture physiological complexities. In contrast, younger patients benefit from transfer learning, likely due to more homogeneous physiological responses. These findings underscore the importance of demographic considerations while demonstrating transfer learning’s potential to improve performance across diverse clinical settings.

\begin{figure}[h!]
	\centering
    
	\begin{minipage}[b]{0.5\linewidth}
		\centering
		\includegraphics[width=\linewidth]{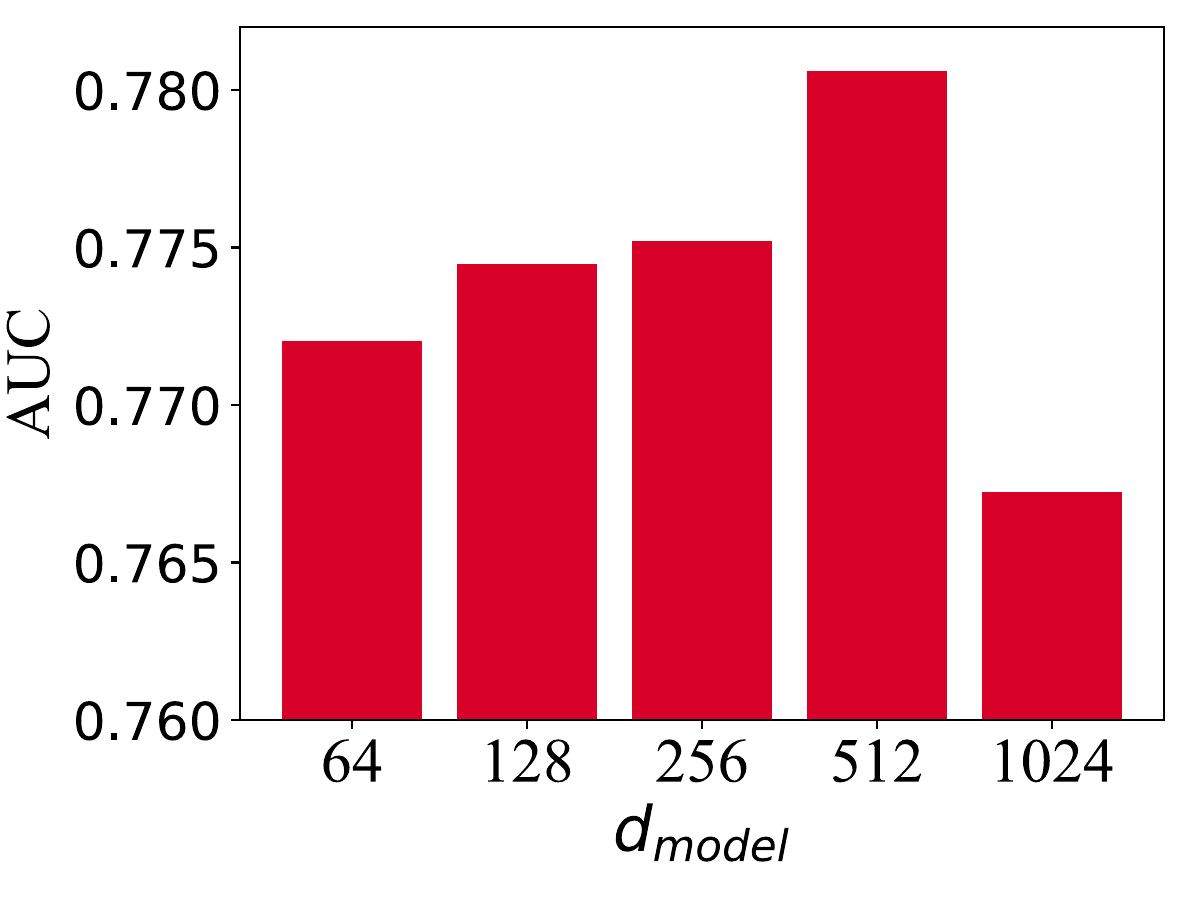}
	\end{minipage}\hfill
	\begin{minipage}[b]{0.5\linewidth}
		\centering
		\includegraphics[width=\linewidth]{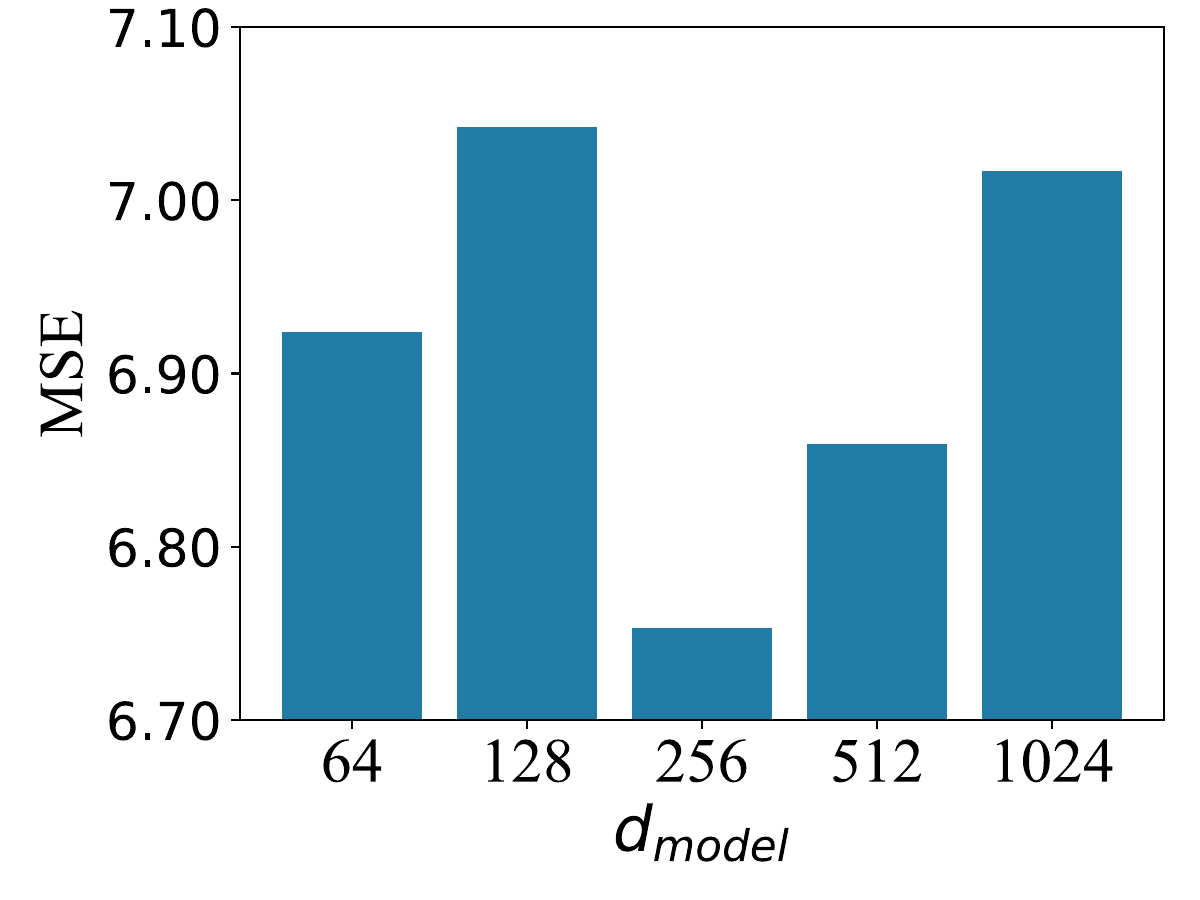}
	\end{minipage}
	\caption{Analysis of MSE and AUC performance for IOH segments across varying $d_{\text{model}} \in \{64, 128, 256, 512, 1024\}$.}
	\label{fig1ablation}
\end{figure}

\begin{figure}[h!]
	\centering
	\begin{minipage}[b]{0.5\linewidth}
		\centering
		\includegraphics[width=\linewidth]{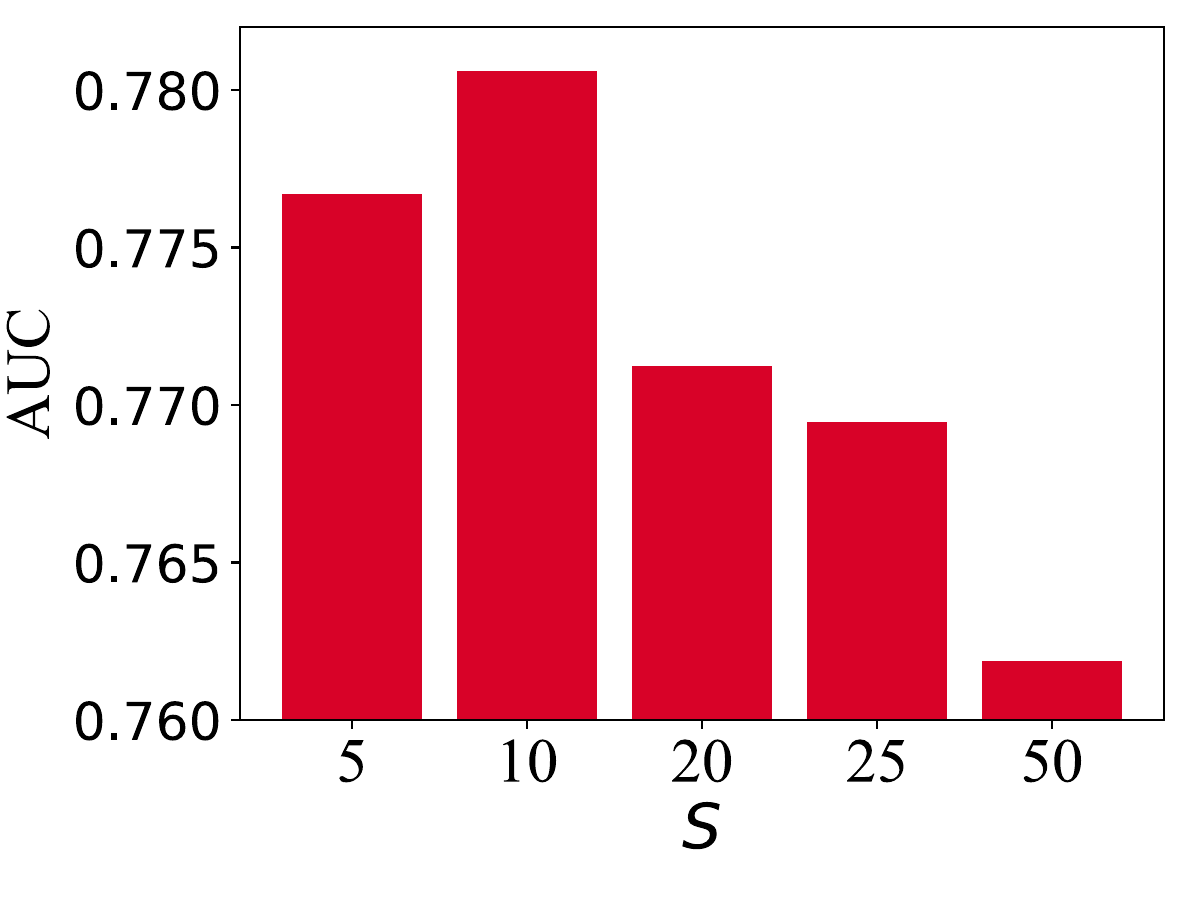}
	\end{minipage}\hfill
	\begin{minipage}[b]{0.5\linewidth}
		\centering
		\includegraphics[width=\linewidth]{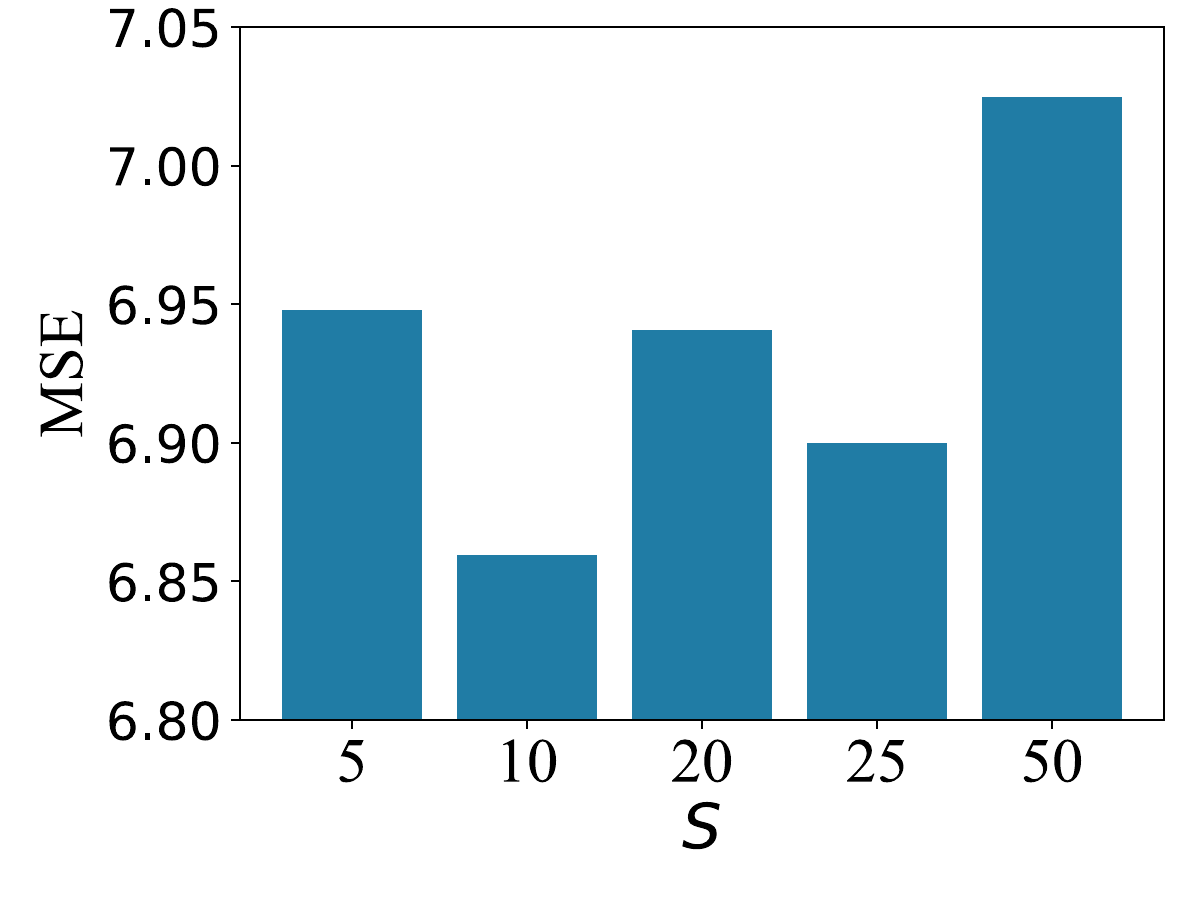}
	\end{minipage}
	\caption{Analysis of MSE and AUC performance for IOH segments across varying \(S\) \(\in \{5, 10, 20, 25, 50\}\).}
	\label{fig2ablation}
\end{figure}

\paragraph{Parameter Sensitivity Analysis.}

To optimize key parameters and enhance the Patch Encoder’s ability to capture temporal dependencies, we perform sensitivity analysis on the CH-OPBP dataset, focusing on model dimension \(d_{\text{model}}\) and patch length \(S\). We first vary \(d_{\text{model}}\) with a fixed context window of 300 and prediction length of 100. As shown in Figure~\ref{fig1ablation}, performance is sensitive to \(d_{\text{model}}\), with a clear optimum maximizing AUC. We then examine the effect of varying \(S\) while keeping \(d_{\text{model}}\) constant. Since \(S\) governs the model’s ability to capture local temporal patterns, its selection is important. Results in Figure~\ref{fig2ablation} confirm that appropriate \(S\) values significantly affect forecasting accuracy. These findings guide the selection of optimal hyperparameters to ensure performance and highlight the importance of configuring the Patch Encoder to effectively model blood pressure dynamics.



\section{Conclusion}

In this paper, we proposed a novel approach to intraoperative hypotension (IOH) prediction by reformulating it as a dynamic sequence forecasting task, enabling proactive and continuous risk assessment. To address key challenges inherent in modeling arterial blood pressure dynamics, we introduced symmetric normalization to mitigate non-stationarity in mean arterial pressure (MAP) and systolic blood pressure (SBP) series, sequence decomposition to better capture complex temporal patterns, and a patch-based Transformer model to enhance representation learning and forecasting accuracy. Extensive experiments on real-world intraoperative datasets validated the effectiveness and robustness of our HMF framework, demonstrating its strong performance in improving early IOH forecasting under various clinical conditions.

\clearpage

\section*{Acknowledgements}

This research was supported by grants from the grants of Provincial Natural Science Foundation of Anhui Province (No.2408085QF193), USTC Research Funds of the Double First-Class Initiative (No. YD2150002501), the Fundamental Research Funds for the Central Universities of China(Grant No. PA2024GDSK0112, No. WK2150110032), Anhui Provincial Natural Science Foundation (No. 2308085MG226), the Key Technologies R\&D Program of Anhui Province (No. 202423k09020039).

\bibliographystyle{named}
\bibliography{ijcai25}

\end{document}